\documentclass{article} 
\usepackage{iclr2021_conference,times}

\usepackage{amsmath,amsfonts,bm}

\def\eqref#1{equation~\ref{#1}}

\def\1{\bm{1}}

\DeclareMathAlphabet{\mathsfit}{\encodingdefault}{\sfdefault}{m}{sl}
\SetMathAlphabet{\mathsfit}{bold}{\encodingdefault}{\sfdefault}{bx}{n}

\usepackage{hyperref}
\usepackage{url}
\usepackage{mwe}
\usepackage{svg}
\usepackage{float}
\usepackage{subfigure}

\title{Practical Deepfake Detection: \\ Vulnerabilities in Global Contexts} 

\iclrfinalcopy

\author{Yang A. Chuming\thanks{Equal contribution.}, \ Daniel J. Wu\footnotemark[1], \ Ken Hong\footnotemark[1]\\
Department of Computer Science\\
Stanford University\\
\texttt{\{ayang, danjwu, kenhong\}@cs.stanford.edu} \\
}

\begin{document}

\maketitle

\begin{abstract}

Recent advances in deep learning have enabled realistic digital alterations to videos, known as deepfakes. This technology raises important societal concerns regarding disinformation and authenticity, galvanizing the development of numerous deepfake detection algorithms. At the same time, there are significant differences between training data and in-the-wild video data, which may undermine their practical efficacy. We simulate data corruption techniques and examine the performance of a state-of-the-art deepfake detection algorithm on corrupted variants of the FaceForensics++ dataset.
While deepfake detection models are robust against video corruptions that align with training-time augmentations, we find that they remain vulnerable to video corruptions that simulate decreases in video quality. Indeed, in the controversial case of the video of Gabonese President Bongo's new year address, the algorithm, which confidently authenticates the original video, judges highly corrupted variants of the video to be fake. Our work opens up both technical and ethical avenues of exploration into practical deepfake detection in global contexts.
\end{abstract}

\vspace{-1em}

\section{Motivation} 
\vspace{-0.5em}

In an era of widespread disinformation, video proof carries a sense of indisputability. However, with the advent of deepfake technology, computer-generated fake videos have become increasingly convincing, subverting the medium's natural trustworthiness \citep{day2019future}. 
In regions with less-developed technology infrastructure, Internet is slow, unreliable, and bandwidth-constrained \citep{itu2019measuring}. Residents of these areas may only have access to video streams which are heavily compressed, artifact-laced, and low-resolution. Furthermore, as residents of these areas tend to have lower levels of technological literacy \citep{literacy_2017}, it is likely that they are less acquainted with deepfake technology. Such populations are highly vulnerable to deception with deepfake videos. 

Many technological solutions have been proposed to tackle deepfake detection \citep{sabir2019recurrent, guera2018deepfake, amerini2019deepfake}. However, these methods are generally trained and tested on videos with high quality and fidelity -- assumptions which are unlikely to hold in the real-world, particularly in areas with less-developed technology infrastructure. In this work, we build a video corruption pipeline, and demonstrate that an existing, state-of-the-art video authentication model is insufficient when applied to lower-quality videos. 
\begin{figure}[b]
    \centering
    \subfigure[]{\includegraphics[width=0.3\textwidth]{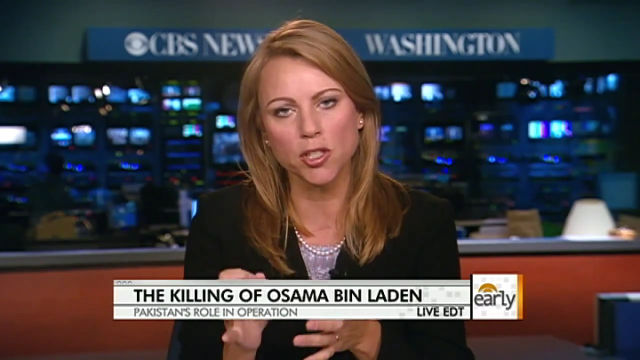}}\quad
    \subfigure[]{\includegraphics[width=0.3\textwidth]{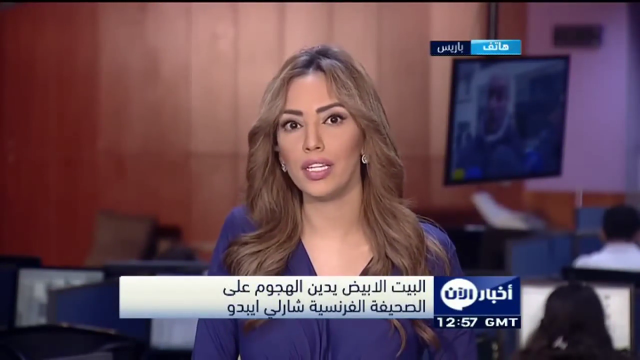}}\quad
    \subfigure[]{\includegraphics[width=0.3\textwidth]{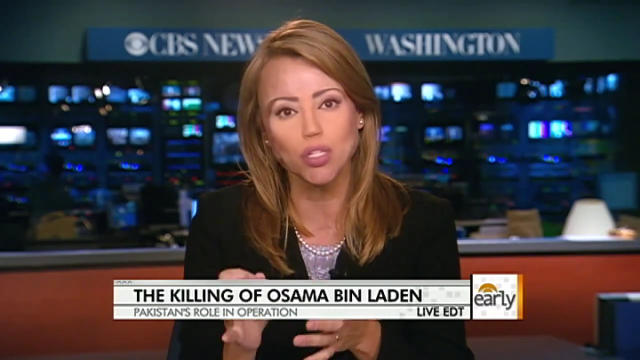}}
    \caption{An example of a synthesized deepfake from the FaceForensics++ dataset. The deepfake (c) is generated by placing the face from the source video (b) onto the target video (a).}
    \label{fig:deepfake_example}
\end{figure}

\paragraph{Our contributions:} In this paper, we (i) propose a video corruption pipeline to simulate quality degradation by applying several video transformations, (ii) apply our corruption pipeline to both legitimate and fake videos, and assess the performance of a state-of-the-art deepfake detection model on our corrupted videos, (iii) identify the vulnerability of video authentication models to corruptions which do not align with training-time data augmentations, and (iv) explore the broader ethical implications of this vulnerability, taking the infamous Gabonese presidential address video as a case study.

\section{Dataset}
\vspace{-0.5em}

We conduct our experiments on the FaceForensics++ dataset \citep{faceforensics}. The dataset is composed of 1000 legitimate videos of news anchors and a corresponding set of 1000 deepfakes. Each video has its audio removed and is clipped to a short 5 to 15 second segment, wherein the anchor is the primary focus of the video. Each deepfake was generated by taking a legitimate video, and applying FaceSwap \citep{faceswap} with the face from another video (Figure \ref{fig:deepfake_example}).

\section{Data corruption methods}
\begin{figure}
    \centering
    \includegraphics[width=\linewidth]{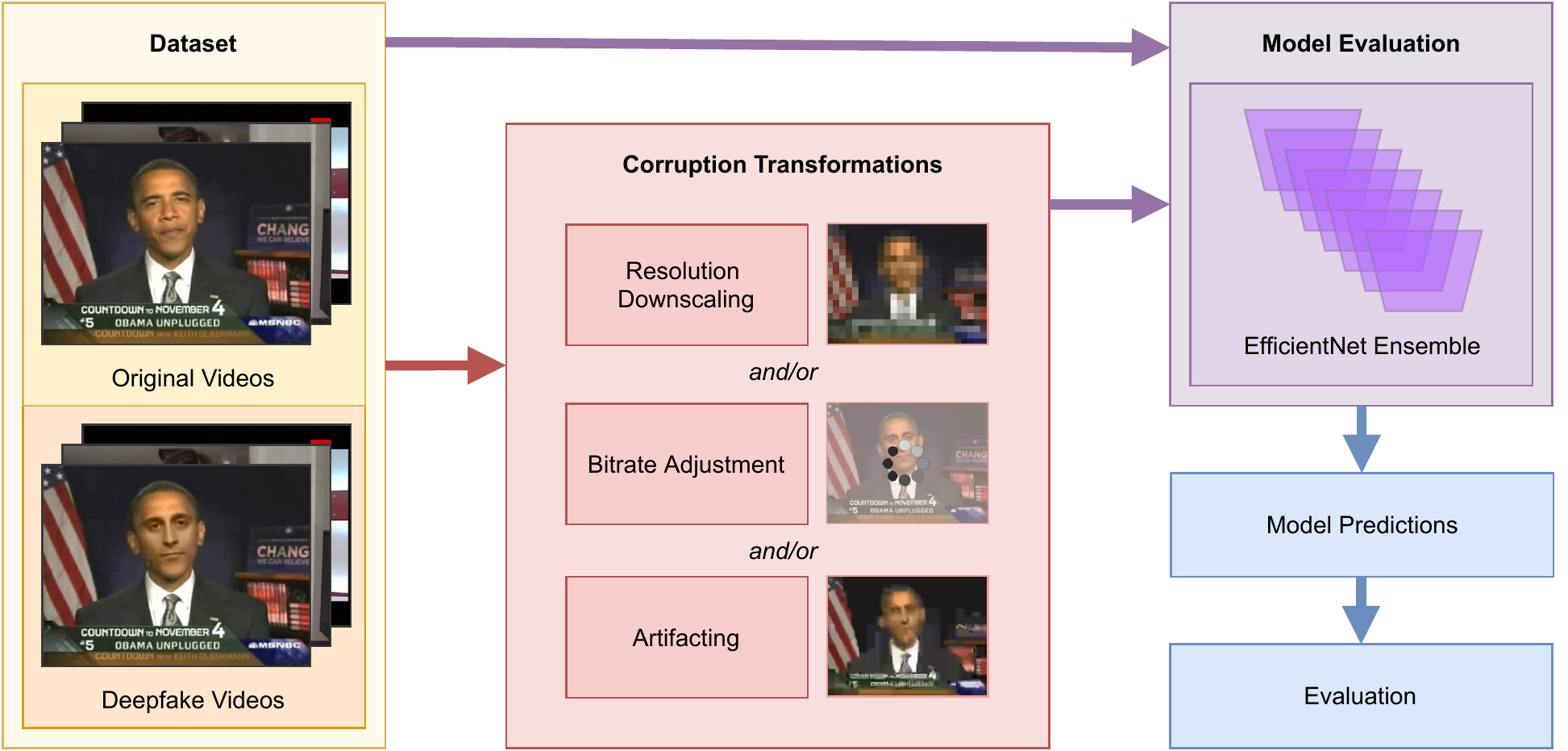}
    \caption{An overview of our method. Both legitimate videos and their corresponding deepfakes are passed through our corruption pipeline, and the resultant videos are scored by a state-of-the-art deepfake detection model.}
    \label{fig:method}
\end{figure}

\vspace{-0.5em}
We propose video corruption methods that result in new videos that are visually similar to those that are organically corrupted (Figure \ref{fig:method}). 

\paragraph{Bitrate:}
Video bitrate refers to the continuous rate in bits at which a video is a digitally represented. 
In areas with low bandwidth, videos are subject to lower bitrates \citep{bitrate_developingcountries}
, generally resulting in lower quality video \citep{bitrate}. For instance, 720p videos on YouTube range from 1.5 to 4 megabits per second (Mbit/s), while 240p videos range from 0.3 to 0.7 Mbit/s \citep{youtubebitrate}. We lower the bitrates of videos to approximately 1 Mbit/s and 0.5 Mbit/s. Figure \ref{fig:bitrate examples} in the Appendix displays frames from the deepfake video after lowering the video bitrate to 1 Mbit/s and 0.5 Mbit/s.

\paragraph{Resolution:}
Video resolution greatly affects the clarity of frames in a video. A resolution of 480p refers to video with vertical height of 480 pixels. Figure \ref{fig:resolution_examples} in the Appendix displays frames from the deepfake video at 720p, 480p, and 240p resolutions.

\paragraph{Constant rate factor (CRF):}
The CRF is a quality control setting, ranging from 0 (best quality) to 51 (worst quality). Videos with lower CRFs (and thus better quality) appear sharper and less blurry than videos with higher CRFs \citep{crf}. We set CRFs of videos at 30, 40, and 50 (Appendix, Figure \ref{fig:crf_examples}). We note that, unlike controlling the bitrate to a constant number, adjusting the CRF leads to a variable bitrate that achieves a specified video quality level.

\paragraph{Datamosh:}
Datamoshing is a glitching technique where only the ``moving parts'' of a video update, thereby creating a pixelated, distorted effect \citep{datamosh}. Figure \ref{fig:datamosh_examples} in the Appendix displays the deepfake video frame after applying the datamoshing effect.

\paragraph{Combination effects:}

In addition to corrupting videos using each of the methods above in isolation, we also corrupt videos by applying the methods sequentially. We separate out our corruption regimes into two tracks: resolution downsampling $\rightarrow$ bitrate reduction; and resolution downsampling $\rightarrow$ CRF increase $\rightarrow$ datamosh. The former track is representative of naturally-occuring video corruption, while the latter is representative of inorganic video corruption.

\section{Model}
\vspace{-0.5em}

In order to provide a faithful replication of state-of-the-art deepfake detection models, we use the top-scoring model in the 2020 Kaggle Deepfake Detection Challenge \citep{DFDCgithub}. This model is an ensemble of 7 EfficientNet B7s \citep{tan2019efficientnet} trained on the Deepfake Detection Challenge Dataset. This dataset is disjoint from the FaceForensics++ dataset, and so we also capture the realistic performance of this model on out-of-domain data.

This model is fairly compute-intensive, containing a total of 462 million parameters and requiring 259 billion FLOPs in each forward pass. Eventual production models, particularly if limited by the low-resource environments in developing countries, may choose to trade accuracy for speed, and thus have poorer performance. As such, we consider our results to be an upper bound on discriminator performance.
\vspace{-0.5em}

\section{Results}
\vspace{-0.5em}

\begin{table}
\parbox{.45\linewidth}{
\centering
\begin{tabular}{|l|c|}
\hline
\multicolumn{1}{|c|}{\bfseries Corruptions} & \multicolumn{1}{|c|}{\bfseries Accuracy}  \\ \hline\hline
720p                  & 86.6 \% \\ 
480p                  & 89.1\% \\ 
240p                  & 90.8\% \\ \hline
BR1.0                   & 88.7\% \\ 
BR0.5                  & 89.1\% \\ 
CRF30                 & 90.3\% \\ 
CRF40                 & 91.4\% \\ 
CRF50                 & 96.7\%\\ \hline
720p + BR1.0             & 87.4\% \\
720p + BR0.5            & 87.7\%\\
720p + CRF30           & 87.9\%\\
720p + CRF30 + Datamosh & 80.8\% \\ \hline
480p + BR1.0             & 88.7\% \\
480p + BR0.5            & 89.2\% \\
480p + CRF40           & 92.6\% \\
480p + CRF40 + Datamosh & 88.2\%\\ \hline
240p + BR1.0             & 92.8\%\\
240p + BR0.5            & 91.3\%\\
240p + CRF50           & 68.5\% \\
240p + CRF50 + Datamosh & \textbf{55.6\%} \\ \hline \hline
\textbf{Uncorrupted videos}              & \textbf{88.9\%} \\ \hline
\end{tabular}
\caption{Video authentication accuracy: The percentage of legitimate videos that the model correctly identified as legitimate, under a variety of corruption schemes.}
}
\hfill
\parbox{.45\linewidth}{
\centering
\begin{tabular}{|l|c|}

\hline
\multicolumn{1}{|c|}{\bfseries Corruptions} & \multicolumn{1}{|c|}{\bfseries Accuracy}  \\ \hline\hline
720p                  & 95.9\%\\ 
480p                  & 93.6\%\\ 
240p                  & 85.1\%\\ \hline 
BR1.0                   & 95.6\% \\ 
BR0.5                  & 92.7\% \\ 
CRF30                 & 89.9\% \\ 
CRF40                 & 73.2\% \\ 
CRF50                 & 26.2\% \\ \hline 
720p + BR1.0             & 96.0\%  \\ 
720p + BR0.5            & 92.4\% \\ 
720p + CRF30           & 91.7\% \\ 
720p + CRF30 + Datamosh & 91.1 \%\\ \hline
480p + BR1.0             & 95.8\% \\ 
480p + BR0.5            & 94.0\%  \\ 
480p + CRF40           & 70.4\% \\
480p + CRF40 + Datamosh & 71.1\% \\ \hline 
240p + BR1.0             & 88.0\% \\ 
240p + BR0.5            & 87.3\% \\ 
240p + CRF50           & \textbf{58.6\%} \\ 
240p + CRF50 + Datamosh & 69.0\% \\ \hline \hline
\textbf{Uncorrupted deepfakes}           & \textbf{95.5\%} \\ \hline 
\end{tabular}
\caption{Deepfake detection accuracy: The percentage of deepfake videos that the model correctly identified as fake, under a variety of corruption schemes.}
}
\end{table}

We assessed our deepfake detection model on FaceForensics++ under a variety of corruption regimes, and our results are presented in Tables 1 and 2. It is important to consider how machine learning practitioners can ensure models are robust to the kinds of corruptions we present. An immediate parallel can be drawn between corruptions caused by poor infrastructure and the purposeful corruptions introduced by data augmentation during training. Indeed, conventional computer vision models are often trained on highly augmented data. These augmentations may be similar to the corruptions that we explore -- for instance, downsampling is a direct analog to decreasing video resolution, and introducing Gaussian noise may be similar to the effects of bitrate reduction. In particular, the model we assessed was trained with Gaussian noise, Gaussian blurring, random horizontal flipping, isotropic resizing, grayscaling, shifting and scaling, FancyPCA, random contrast sharpening and brightness variation, and hue saturation shifts. 

We note that the most significant degradation in performance occurs due to high constant rate factor and datamoshing. In particular, CRF50 in the absence of other corruptions reliably fools the model into authenticating deepfakes. This is telling, as these corruptions do not have a direct analogue to a train-time augmentation, and they dramatically alter the RGB values of the video while remaining semantically cognizable to human observers.

\section{Gabon: A Case Study}
\vspace{-0.5em}

One of the most infamous occurrences of a suspected deepfake occurred in the nation of Gabon \citep{motherjonesgabon, washpostgabon, westerlund2019emergence, jones2020artificial}. After President Ali Bongo Ondimba was hospitalized in 2018, his lack of public appearances over the next several months fueled rumors that he was incapacitated. On New Year's Day, 2019, the Gabonese government released a \emph{legitimate} video of President Bongo delivering a public address. This video was widely perceived to be fake, motivating members of the Gabonese military to attempt a coup just a week later \citep{motherjonesgabon}.

At the time, deepfake detection algorithms had classified the video as legitimate with high confidence \citep{washpostgabon}, and our model was consistent with this finding. Our model (which produces scores close to 1 if it identifies a deepfake and 0 otherwise) gave the video a score of only 0.0119. However, we found that after applying strong corruptions, which emulates the experience of viewing this video in areas with poor Internet access, e.g. Gabon \citep{internetspeed}, the model confidence changed dramatically. When the video was downsampled to 240p and re-encoded with a CRF of 50, the model produced a score of 0.565, which causes the model to incorrectly label the video as a deepfake. This demonstrates corrupted videos may `fool' deepfake detection systems and create dangerous sociopolitical implications.

\section{Discussion}
\vspace{-0.5em}

There are a multitude of ethical risks which arise from the combination of deepfake technology and video corruption, especially in areas with developing technology infrastructure. Viewers in developing nations are particularly vulnerable to disinformation. Bandwidth and resource constraints in developing regions may undermine the predictive power of deepfake discriminators, allowing low-quality deepfakes past automated filters. In turn, viewers may have difficulty determining the authenticity of low-quality videos. Moreover, viewers without a considerable degree of technological literacy may not know that certain data corruption methods are inorganic -- malicious actors can target such viewers by applying inorganic corruption methods to fool both the judgments of deepfake discriminators and viewers. Our work serves as a fertile starting ground for further research addressing the potential for manipulating videos to deceive both human viewers and deepfake discriminators. Although we focus on video corruption, previous work in generating adversarial examples, such as that described in \cite{eykholt}, suggests that there are myriad other methods to deceive algorithms beyond simulating changes in video quality. We hope that our work galvanizes the scientific community to closely consider the practical implications, limitations, and vulnerabilities of deepfake detection technologies in a global context.

\newpage

\bibliography{iclr2021_conference}
\bibliographystyle{iclr2021_conference}

\appendix
\section{Appendix}
\vspace{-0.5em}


\begin{figure}[H]
    \centering
    \subfigure[]{\includegraphics[width=0.3\textwidth]{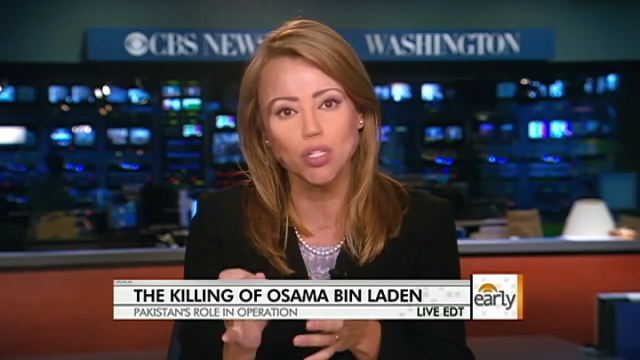}}\quad
    \subfigure[]{\includegraphics[width=0.3\textwidth]{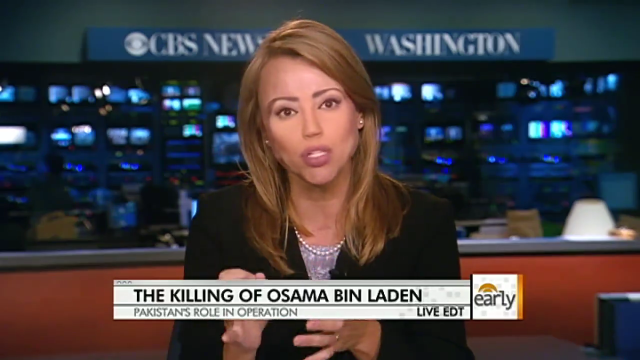}}\quad
    \caption{Deepfake video frames at approximately 1.0 and 0.5 Mbit/s bitrate}
    \label{fig:bitrate examples}
\end{figure}


\begin{figure}[H]
    \centering
    \subfigure[]{\includegraphics[width=0.3\textwidth]{978_975_720p.mp4.png}}\quad
    \subfigure[]{\includegraphics[width=0.3\textwidth]{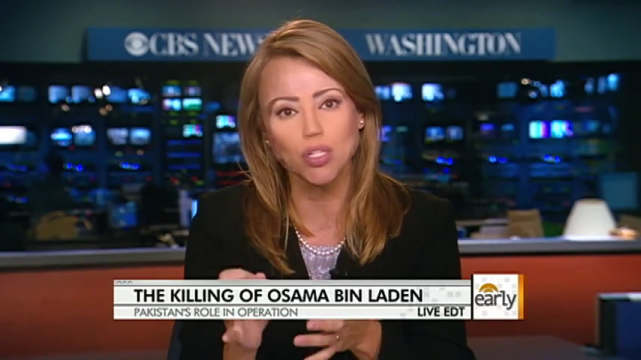}}\quad
    \subfigure[]{\includegraphics[width=0.3\textwidth]{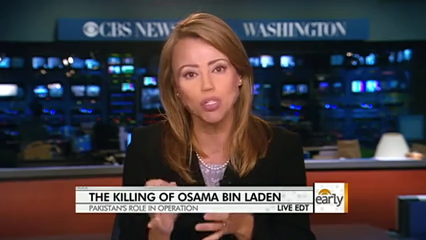}}
    \caption{Deepfake video frames at 720p, 480p, and 240p resolutions}
    \label{fig:resolution_examples}
\end{figure}


\begin{figure}[H]
    \centering
    \subfigure[]{\includegraphics[width=0.3\textwidth]{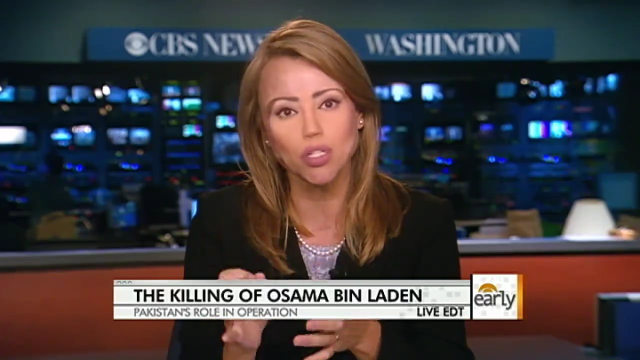}}\quad
    \subfigure[]{\includegraphics[width=0.3\textwidth]{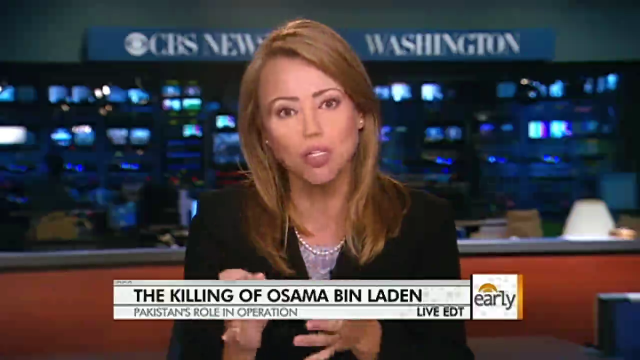}}\quad
    \subfigure[]{\includegraphics[width=0.3\textwidth]{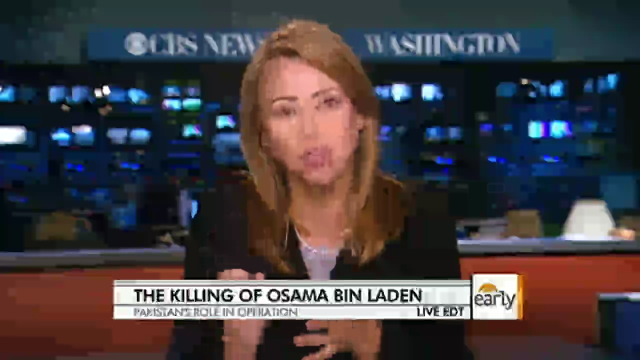}}
    \caption{Deepfake video frames at CRF 30, 40, and 50}
    \label{fig:crf_examples}
\end{figure}


\begin{figure}[H]
    \centering
    \subfigure[]{\includegraphics[width=0.3\textwidth]{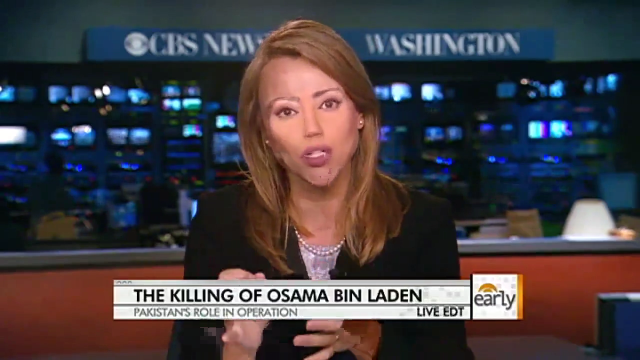}}\quad
    \subfigure[]{\includegraphics[width=0.3\textwidth]{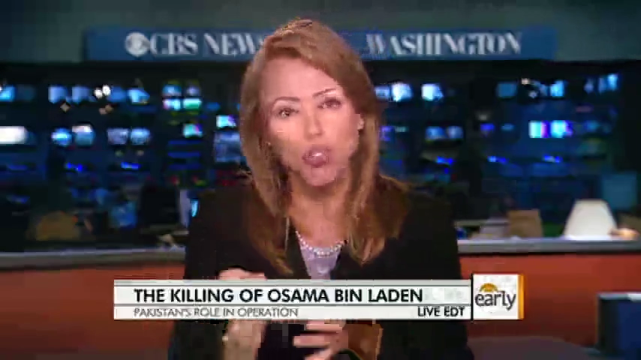}}\quad
    \subfigure[]{\includegraphics[width=0.3\textwidth]{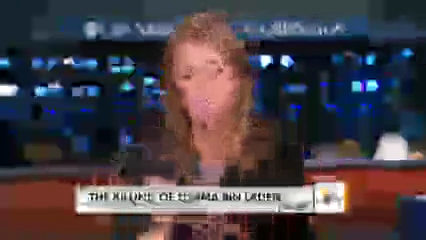}}
    \caption{Deepfake video frames with datamoshing effect: (a) shows datamoshing at 720p and CRF 30; (b) shows datamoshing at 480p and CRF 40; (c) shows datamoshing at 240p and CRF 50}
    \label{fig:datamosh_examples}
\end{figure}


\end{document}